\documentclass[11pt]{article}

\usepackage[preprint]{acl}

\usepackage{times}
\usepackage{latexsym}
\usepackage[T1]{fontenc}
\usepackage[utf8]{inputenc}
\usepackage{microtype}
\usepackage{inconsolata}
\usepackage{graphicx}
\usepackage{pgfplots}
\usepgfplotslibrary{groupplots}
\usetikzlibrary{arrows.meta}

\usepackage{amsmath,amssymb}
\usepackage{wasysym}
\DeclareUnicodeCharacter{266B}{\twonotes}
\usepackage{booktabs}
\usepackage{multirow}
\usepackage{xspace}
\usepackage{algorithm}
\usepackage{algpseudocode}

\graphicspath{{figures/}}

\title{Metacognitive Decoding: Efficient Inference for Diffusion Language Agents}

\author{
  Mingyu Zhao\textsuperscript{1}\thanks{These authors contributed equally to this work.} \quad Shuhang Lin\textsuperscript{1}\footnotemark[1] \quad Ligong Han\textsuperscript{1} \quad Xiaojing Yu\textsuperscript{3} \\
  \textbf{Zhenting Wang\textsuperscript{1} \quad Yifei Pang\textsuperscript{2} \quad Yang Zhou\textsuperscript{1} \quad Dimitris N. Metaxas\textsuperscript{1}} \\
  {\small\textsuperscript{1}\texttt{\{zhao.mingyu,shuhang.lin,zhenting.wang,eta.yang\}@rutgers.edu}} \\
  {\small\textsuperscript{1}\texttt{\{ligong.han,dnm\}@rutgers.edu}} \\
  {\small\textsuperscript{2}\texttt{yifeip@cs.unc.edu}\quad\textsuperscript{3}\texttt{yyxjing@gmail.com}}
}

\begin{document}
\maketitle

\begin{abstract}

Diffusion language models offer a promising path to efficient agents through parallel token generation.
However, efficient interaction requires adapting both where computation is spent and how much text is generated.
We introduce a metacognitive controller that guides diffusion generation according to the needs of the current response and the agent's interaction with the environment.
It learns when to finalize tokens and how far generation should extend, reducing overall computation within responses. Our controller, a lightweight plug-in for the diffusion backbone, is trained on agent trajectories to balance correctness and generation cost. This approach connects the agent's generation process to the interaction the agent serves. Experiments across five benchmarks demonstrate that our approach reduces inference FLOPs by 20.6\% on average compared with the most efficient baseline, while maintaining competitive task accuracy.
Ablation studies further validate the effectiveness of our approach.

\end{abstract}

\section{Introduction}
\label{sec:intro}

\begin{figure*}[t]
  \centering
  \includegraphics[width=\textwidth]{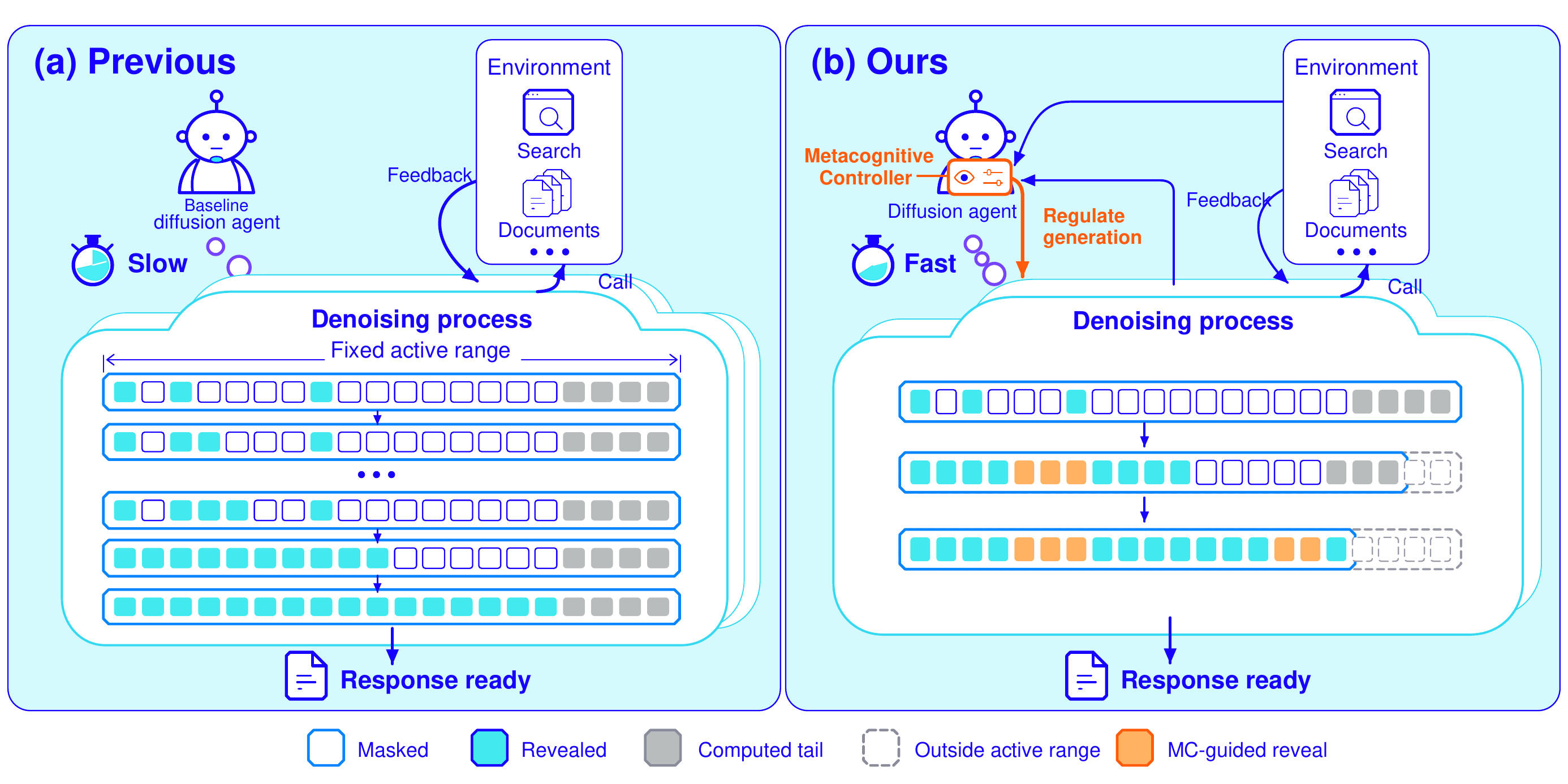}
  \caption{Metacognitive control of environment-aware diffusion decoding.
  Left: fixed-range decoding computes positions beyond the response consumed by the environment.
  Right: our controller guides content commitment and adjusts the active computation range during interaction. 
  Orange denotes controller-guided revelation, gray denotes computed tail positions, and dashed cells lie outside the active range.}
  \label{fig:teaser}
\end{figure*}

Diffusion language models offer a promising foundation for efficient language agents.
Their ability to generate multiple tokens in parallel creates opportunities to accelerate generation~\citep{nie2025llada,wu2025fastdllm}.
In an agent, however, each response is part of an ongoing interaction: the model reasons about a task, calls a tool, observes the result, and continues.
What the agent needs to generate changes throughout this process.
Making effective use of diffusion therefore introduces a new challenge: adapting diffusion generation to the dynamic needs of agentic reasoning and interaction.

The primary bottleneck lies in the mismatch between conventional rigid decoding mechanisms and the highly structured nature of agent trajectories, which leads to severe computational redundancy.
This inefficiency manifests in two main ways.
First, agent outputs contain a mix of highly predictable formatting (e.g., API templates) and complex reasoning. Conventional decoding treats all tokens uniformly; it either wastes compute by over-processing easy tokens or introduces errors by finalizing complex tokens too early.
Second, agent interactions have strict execution boundaries. Once a tool call is generated, the agent must pause for environment feedback. However, because diffusion models process tokens in fixed-size blocks, they often wastefully compute trailing tokens beyond this boundary that are ultimately discarded.
Therefore, to achieve high efficiency, a diffusion agent must dynamically adjust its decoding strategy based on both prediction confidence and interaction states.

To address these inefficiencies, we introduce a \textbf{metacognitive controller} that enables \emph{environment-aware generation} (Figure~\ref{fig:teaser}).\footnote{We release our implementation on \href{https://github.com/IlluminatingFish/metacognitive-controller}{GitHub}.}
Instead of generating tokens blindly, the controller monitors the agent's interaction states and the model's prediction confidence to dynamically coordinate two decisions: which tokens to finalize and how many positions to compute.
At each step, after the diffusion backbone predicts tokens, the controller applies a cascaded policy to optimize decoding:
First, \emph{flux decoding} optimizes the active computation range. It dynamically shrinks the generation window to exclude unnecessary downstream positions from computation, expanding it back only when the current response actually requires more space to complete.
Second, \emph{confidence alignment decoding} manages token acceptance. It identifies highly predictable structures (e.g., standard tool-call formats) and finalizes them simultaneously, while adaptively adjusting the acceptance thresholds for individual tokens.
Together, these mechanisms ensure that computation is spent only where needed, allowing generation to halt efficiently the moment a tool call is triggered.

Our contributions are threefold:
\begin{itemize}
\item We identify a diffusion agent efficiency bottleneck: diffusion agents can have low token efficiency with common patterns.
\item We propose a metacognitive controller for efficient agent generation. Our controller monitors both model prediction signals and agent interactions. It controls agent decoding with a cascaded decoding policy: confidence alignment decoding controls structural segments and token acceptance, while flux decoding adjusts the computation range through recoverable empty tokens.
\item Across five benchmarks, our approach outperforms existing baselines by reducing inference FLOPs while maintaining competitive task accuracy. Ablation studies further demonstrate the effectiveness of our metacognitive controller.
\end{itemize}

\section{Related Work}
\label{sec:related}

\paragraph{Diffusion language models.}
Diffusion language models generate text through iterative denoising, allowing multiple token positions to be predicted in parallel rather than following a fixed left-to-right order.
Early work explored diffusion in continuous embedding spaces~\citep{li2022diffusionlm} and discrete token spaces~\citep{austin2021structured}.
Subsequent work introduced score entropy for discrete diffusion~\citep{lou2024sedd} and simplified masked diffusion training to a weighted masked language modeling objective~\citep{sahoo2024masked}.
Building on these advances, recent work has scaled diffusion language models through pre-training from scratch, as in LLaDA~\citep{nie2025llada}, and adaptation from pretrained autoregressive models~\citep{gong2025scaling}.
In particular, LLaDA demonstrates instruction-following, in-context learning, and reasoning capabilities at the 8B scale~\citep{nie2025llada}.
Beyond model scaling, block diffusion extends diffusion generation to variable-length sequences by combining autoregressive generation across blocks with parallel denoising within each block~\citep{arriola2025block,llada2}.

\paragraph{Language agents and tool use.}
Beyond standalone text generation, language agents use external tools and environmental feedback to solve tasks.
ReAct interleaves reasoning with actions, allowing observations to inform subsequent decisions~\citep{yao2023react}, while Toolformer learns when and how to invoke APIs and incorporate their results into generation~\citep{schick2023toolformer}.
Reflexion uses feedback to form verbal reflections that guide subsequent attempts~\citep{shinn2023reflexion}.
Diffusion models have also been explored for interaction and planning.
DiffTOD formulates target-oriented dialogue planning as conditional trajectory generation, using diffusion to construct plans non-sequentially~\citep{du2025difftod}.
More directly, DLLM-Searcher adapts diffusion language models to search agents through supervised fine-tuning and reinforcement learning~\citep{dllmsearcher}.
Its P-ReAct decoding pre-fills tool-call boundaries and biases confidence scores within the tool region to prioritize tool-call completion, allowing tool execution to overlap with reasoning generation.

\paragraph{Adaptive decoding for diffusion language models.}
Decoding policies determine how diffusion language models translate parallel token predictions into efficient generation. Revealing multiple tokens per step reduces the number of model evaluations but can compromise generation quality. Confidence-ranked decoding selects a prescribed number of predictions at each step~\citep{nie2025llada}, whereas confidence-thresholding reveals tokens whose confidence exceeds a specified threshold~\citep{wu2025fastdllm}. Further approaches employ entropy constraints to determine how many tokens
to reveal~\citep{benhamu2025ebsampler}, or use token predictions from
previous denoising steps to determine which tokens to
reveal~\citep{wei2025slowfast,wang2025creditdecoding}. Existing studies have also explored learning which positions to
denoise~\citep{liu2025ddpd} and revising previously generated tokens
by masking them again~\citep{wang2025remdm}. Another line of work focuses on reducing the computational cost of each
denoising step through adaptive caching~\citep{ma2025dkvcache,liu2025dllmcache,nguyentri2025elasticcache}
and sparse attention~\citep{song2025sparsedllm,long2026focusdllm}.

\section{Methodology}
\label{sec:method}

In this section, we present a metacognitive controller for accelerating diffusion agents. It selects which tokens to finalize and how much of the current block to keep active. We train the controller to balance answer correctness and inference cost.
We first describe block diffusion (\S\ref{sec:method-block}), then
introduce the controller (\S\ref{sec:method-structure}) and cascaded
decoding (\S\ref{sec:method-policy}), and finally present controller
training (\S\ref{sec:method-training}).

\subsection{Backbone: Block Diffusion}
\label{sec:method-block}

Diffusion models such as LLaDA2.0~\citep{llada2} generate responses block by block. For example, given the query and current interaction history $\mathbf{c}$, the probability
of a response $\mathbf{x}$ comprising $K$ blocks factorizes as
\begin{equation}
  p(\mathbf{x}\mid\mathbf{c})
  = \prod_{k=1}^{K}
    p(\mathbf{x}^{(k)}
    \mid\mathbf{c},\mathbf{x}^{(<k)}\bigr),
  \label{eq:block}
\end{equation}
where $\mathbf{x}^{(k)}$ denotes
the $k$-th block, and $\mathbf{x}^{(<k)}$ the preceding blocks.
Each block is generated by iterative denoising from masked tokens,
conditioned on completed blocks.

\subsection{Metacognitive Controller}
\label{sec:method-structure}

\begin{figure*}[t]
  \centering
  \includegraphics[width=\textwidth]{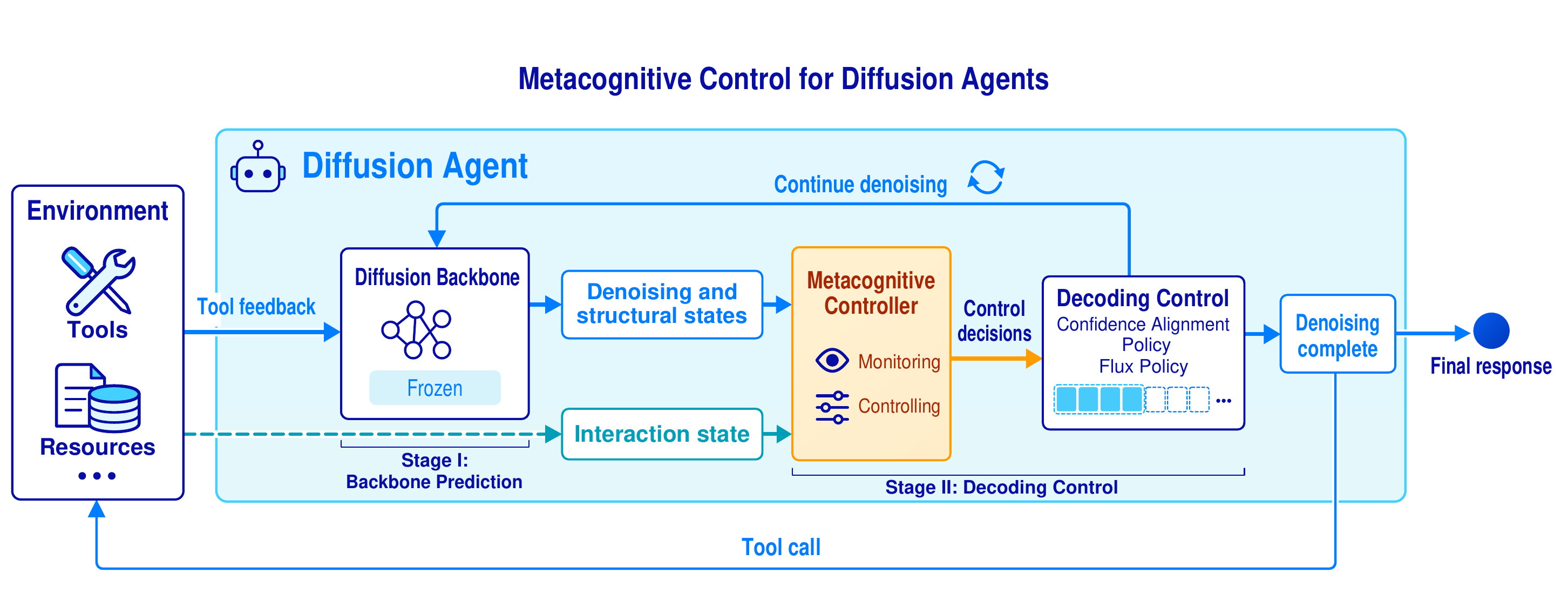}
  \caption{Overview of our approach.
    The metacognitive controller monitors the generation state and
    interaction progress, then guides backbone predictions through
    the confidence alignment policy and flux policy.
    Tool calls provide environmental feedback for subsequent generation.
  }
  \label{fig:shared-controller}
\end{figure*}

In this section, we introduce a metacognitive controller, a lightweight plug-in that assists diffusion agent generation. In general, the controller has two functions: \textit{monitoring} signals from both the model and the external environment, and \textit{controlling} diffusion agent generation.
Figure~\ref{fig:shared-controller} illustrates the workflow of the metacognitive controller.

\paragraph{Monitoring.}
The controller monitors both internal model signals and external interaction signals.
Internal model signals comprise the denoising state and structural state.
The \emph{denoising state} captures prediction uncertainty, token
revelation status, and progress within the current block.
The \emph{structural state} captures recurring format patterns in tool
calls, thinking, and final answers, as well as their compatibility with
the partially generated response.
External interaction signals are represented by the
\emph{interaction state}, which tracks the current interaction turn and
whether thinking segments, tool calls, and final answers have been completed.
These monitored signals are organized into position, structural-candidate,
and global inputs. The controller jointly encodes these inputs to obtain
the corresponding representations for decoding control.

\paragraph{Controlling.}

Our metacognitive controller facilitates generation by diffusion agents. We implement the control mechanism through a two-stage decoding policy. In the first stage, the diffusion backbone produces token predictions.
In the second stage, the metacognitive controller refines the decoding
process through two policies: the \emph{confidence alignment policy}
and the \emph{flux policy}. 
For \emph{flux decoding}, it determines which positions remain active
and which are excluded from computation until recovery. For \emph{confidence alignment decoding}, it scores structural candidates derived from the recurring format patterns and predicts acceptance thresholds for both structural candidates and individual tokens. We describe these policies in detail in Section~\ref{sec:method-policy}.

\subsection{Cascaded Decoding Policy}
\label{sec:method-policy}

Our decoding design follows two principles. First, we keep the diffusion backbone
frozen to avoid costly backbone post-training, while leveraging the
monitored signals to guide generation. Second, we seek to minimize
interference with the backbone's original generation. To apply these principles, we propose a cascaded decoding policy. It contains two layers: the diffusion backbone first produces token predictions as usual; the controller then uses these predictions and the monitored signals to guide decoding through the \emph{confidence alignment policy} and \emph{flux policy}.

\subsubsection*{Stage I: Backbone Prediction}

At each denoising step, the diffusion backbone follows its standard
prediction procedure, producing a token distribution at each masked
position in the current block, conditioned on the interaction history,
completed blocks, and the current block state.

Let $p_i(v)$ denote the predicted probability of token $v\in\mathcal V$
at position $i$, where $\mathcal V$ is the backbone vocabulary.
The predicted token and its confidence are
\begin{equation}
  \hat{x}_i=\arg\max_{v\in\mathcal V}p_i(v),
  \qquad c_i=p_i(\hat{x}_i).
  \label{eq:backbone-prediction}
\end{equation}
These predictions are passed to the second-stage controller to determine
which tokens are accepted and which positions remain active.

\subsubsection*{Stage II: Decoding Control}

\paragraph{Confidence Alignment Policy.}

During agent decoding, redundant patterns recur in generated responses.
For example, tool-use agents repeatedly generate structured tool-call
formats, while question-answering agents reuse formats for thinking
and final answers.
Therefore, to exploit this redundancy, we introduce the
\emph{confidence alignment policy}. Specifically, we extract recurring structural segments from sampled trajectories
of the agent task to construct structural candidates. During decoding,
each candidate pairs a segment with a compatible location in the
current block. We retain only structural candidates that match the already generated tokens at their proposed locations.

Specifically, let $J$ denote the index set of these candidates. For each candidate $j\in J$, the controller uses its representation
$\mathbf h_j^{\mathrm{cand}}$ to compute a score $u_j$, indicating its
preference for accepting that segment:
\begin{equation}
  u_j=\mathbf w_c^\top\mathbf h_j^{\mathrm{cand}}+b_c,
  \qquad j^*=\arg\max_{j\in J}u_j.
  \label{eq:draft-parallel}
\end{equation}
Here, $\mathbf w_c$ and $b_c$ are learned parameters, and $j^*$ indexes
the highest-scoring candidate.

The controller predicts a relative acceptance threshold $\rho$ from
the global representation. If $u_{j^*}>\rho$, the selected candidate
is accepted, jointly filling its masked positions with the segment tokens.
For each remaining masked position $i$, the controller predicts a
threshold offset $\delta_i$ from the position representation and
reveals features. The backbone's predicted token $\hat{x}_i$ is accepted
when its confidence $c_i$ satisfies $c_i\geq\tau_0-\delta_i$,
where $\tau_0$ is the base threshold.
These acceptance criteria adapt structural and token-level revelation
to the current decoding state.

\paragraph{Flux Policy.}
We introduce a dynamic empty token $\varepsilon$ to mark a recoverable
endpoint of the active computation range. At each denoising step, we update its position by selecting the highest-probability position predicted by the metacognitive controller. Positions beyond this endpoint do not participate in subsequent denoising.

For a block of original size $B$, the controller uses the position
representations $\mathbf h_i^{\mathrm{pos}}$ to score the possible
lengths of the active computation range:
\begin{equation}
  \begin{aligned}
    s_i &= \mathbf w_r^\top\mathbf h_i^{\mathrm{pos}}+b_r,
       \quad i=1,\ldots,B,\\
    \mathbf p^{\mathrm{range}}
       &=\operatorname{softmax}(s_1,\ldots,s_B).
  \end{aligned}
  \label{eq:draft-range}
\end{equation}
where $\mathbf p^{\mathrm{range}}$ assigns probabilities to different
lengths of the active computation range.
At inference, the decoder selects the highest-probability length that
preserves all committed non-padding tokens. Selecting an active computation
range of length $\ell$ assigns the empty token $\varepsilon$ to position
$\ell$. Positions beyond $\ell$ do not participate in subsequent denoising.
If the active computation range is fully decoded but the turn remains
incomplete, the controller uses the updated decoding state to select
a longer range $\ell'>\ell$, moves the empty token to position $\ell'$,
and restores the additional positions as masks.
Once recovery begins, the active computation range can expand but
cannot shrink within the current block.

\subsection{Controller Training}
\label{sec:method-training}

In this section, we present the training procedure for the metacognitive
controller, which learns a cascaded decoding policy to balance task
correctness and inference cost over complete interaction trajectories.

\paragraph{Trajectory-level reward.}
We reward correct trajectories according to their inference cost.
For a complete trajectory $\tau$ generated for question $q$, the reward is
\begin{equation}
  R(\tau)=
  \begin{cases}
    \left[\max\!\left(0,1-\dfrac{e(\tau)}{\eta}\right)\right]^\alpha,
      & \text{if correct},\\[4pt]
    0, & \text{otherwise}.
  \end{cases}
  \label{eq:reward}
\end{equation}
We define the relative inference cost as
$e(\tau)=F(\tau)/F_{\mathrm{ref}}(q)$,
where $F(\tau)$ is the backbone and controller inference cost, and
$F_{\mathrm{ref}}(q)$ is the inference cost of the same frozen backbone
without the controller on question $q$.
Reference costs are precomputed and held fixed throughout training.
The parameter $\eta$ sets the cost scale, and $\alpha$ controls the
reward's sensitivity to efficiency.
Correctness is evaluated using the task-specific metric.
The reward becomes zero when $e(\tau)\geq\eta$.

\paragraph{Controller optimization.}
At each denoising step during rollout, we sample a length for the active
computation range according to $\mathbf p^{\mathrm{range}}$.
When the selected length leaves the range unchanged, we sample a
structural candidate and whether to accept it according to the controller's
output probabilities. For each remaining masked position $i$ in the active
computation range, we independently sample whether to accept the
backbone-predicted token. Its acceptance probability increases with the
difference between its confidence $c_i$ and the adjusted threshold
$\tau_0-\delta_i$.

We train the controller on-policy following GRPO~\citep{shao2024deepseekmath}.
In each round, we sample a group of complete interaction trajectories
for each question through the current controller and compute their rewards.
We standardize rewards within each group to obtain trajectory-level
advantages.

\begin{table*}[t]
  \centering
  \small
  \setlength{\tabcolsep}{2pt}
  \renewcommand{\arraystretch}{1.15}
  \begin{tabular*}{\textwidth}{@{\extracolsep{\fill}}l*{10}{c}@{}}
    \toprule
    \multirow{2}{*}{Method}
      & \multicolumn{2}{c}{BFCL Live}
      & \multicolumn{2}{c}{BFCL Non-live}
      & \multicolumn{2}{c}{HotpotQA}
      & \multicolumn{2}{c}{2Wiki}
      & \multicolumn{2}{c}{MuSiQue} \\
    \cmidrule(lr){2-3}\cmidrule(lr){4-5}\cmidrule(lr){6-7}\cmidrule(lr){8-9}\cmidrule(l){10-11}
      & {\fontsize{8}{9.5}\selectfont Acc.\,$\uparrow$} & {\fontsize{8}{9.5}\selectfont FLOPs\,$\downarrow$}
      & {\fontsize{8}{9.5}\selectfont Acc.\,$\uparrow$} & {\fontsize{8}{9.5}\selectfont FLOPs\,$\downarrow$}
      & {\fontsize{8}{9.5}\selectfont Acc.\,$\uparrow$} & {\fontsize{8}{9.5}\selectfont FLOPs\,$\downarrow$}
      & {\fontsize{8}{9.5}\selectfont Acc.\,$\uparrow$} & {\fontsize{8}{9.5}\selectfont FLOPs\,$\downarrow$}
      & {\fontsize{8}{9.5}\selectfont Acc.\,$\uparrow$} & {\fontsize{8}{9.5}\selectfont FLOPs\,$\downarrow$} \\
    \midrule

    Fast-dLLM & 57.46 & 57.58 & 52.66 & 54.98 & 43.20 & 324.16 & 67.70 & 340.49 & \textbf{21.20} & 342.32 \\
    LLaDA-8B-Tools & 50.00 & 2889.56 & 37.27 & 1375.23 & 16.50 & 4310.92 & 22.80 & 4185.11 & 2.20 & 2644.99 \\
    DLLM-Searcher & 40.85 & 547.10 & 16.09 & 433.45 & 37.10 & 1277.25 & 50.40 & 1261.22 & 17.80 & 1237.25 \\
    LLaDA-8B (SFT) & 66.52 & 32.49 & 63.60 & 19.28 & 45.70 & 175.00 & 67.10 & 204.51 & 20.70 & 197.38 \\
    LLaDA2.0-mini & 69.72 & 43.79 & \textbf{79.42} & 14.12 & 36.30 & 142.09 & 41.00 & 199.93 & 10.80 & 168.37 \\
    \midrule
    \textbf{Ours} & \textbf{70.96} & \textbf{18.48} & 65.47 & \textbf{12.51} & \textbf{46.60} & \textbf{131.91} & \textbf{67.90} & \textbf{147.68} & 20.30 & \textbf{142.83} \\
    \bottomrule
  \end{tabular*}
  \caption{Task accuracy and inference cost of our method and baseline methods. FLOPs accumulate model computation over the complete interaction trajectory, including controller computation.}
  \label{tab:backbone-em}
  \label{tab:backbone-transfer}
\end{table*}

\section{Experiments}
\label{sec:experiments}

Our experiments evaluate whether environment-aware decoding reduces
inference cost while maintaining task accuracy.
We compare our method with existing diffusion-based agents and language
models on tool-calling and multi-hop question-answering tasks.
We further examine the contributions of individual control components
through ablations and evaluate the method across diffusion backbones.

\subsection{Experiment Setup}
\label{sec:exp-setup}

\paragraph{Datasets and interaction environments.}
We evaluate our metacognitive controller in diffusion-agent settings. 
We sample BFCL v3~\citep{patil2025bfcl} to evaluate tool-calling capabilities, splitting its
non-live and live subsets into training and test sets at a 1:1 ratio
within each subset, following~\citet{lu2025environmenttuning}.
We sample 1,000 questions from each of HotpotQA~\citep{yang2018hotpotqa},
2WikiMultiHopQA~\citep{ho2020twiki}, and MuSiQue~\citep{trivedi-etal-2022-musique} to evaluate multi-hop
reasoning and tool use, following~\citet{dllmsearcher}.

\paragraph{Baselines.}
We compare our approach with general diffusion models, task-adapted diffusion
models, and diffusion decoding acceleration methods. 
\textbf{LLaDA2.0-mini}~\citep{llada2} is an open-source mixture-of-experts
diffusion language model with block diffusion decoding, included as a
general diffusion baseline.
\textbf{LLaDA-8B-Tools}~\citep{proximile2025lladatools} is a version of
LLaDA-8B-Instruct fine-tuned for tool calling, providing a comparison
with a task-adapted diffusion model.
\textbf{DLLM-Searcher}~\citep{dllmsearcher} adapts diffusion language models
into search agents through agentic post-training and introduces P-ReAct
to overlap tool execution with reasoning. We include it as a
diffusion-agent baseline.
\textbf{Fast-dLLM}~\citep{wu2025fastdllm} is a training-free acceleration
method that enables KV caching and confidence-aware parallel decoding.
We include it to compare controller-based decoding with an existing
diffusion acceleration method.
We evaluate it with the same diffusion backbone weights,
tool interfaces, and generation budgets as our method,
differing only in the decoding and caching mechanisms.
\textbf{LLaDA-8B (SFT)} is based on our LLaDA-8B-Instruct backbone fine-tuned on
tool-use trajectories, with only the metacognitive controller removed.
This comparison isolates the contribution of metacognitive control to
accuracy and inference cost.

\begin{table*}[t]
  \centering
  \small
  \setlength{\tabcolsep}{1.5pt}
  \renewcommand{\arraystretch}{1.3}
  \begin{tabular*}{\textwidth}{@{\extracolsep{\fill}}c*{10}{c}@{}}
    \toprule
    \multirow{2}{*}{Setting}
      & \multicolumn{2}{c}{BFCL Live}
      & \multicolumn{2}{c}{BFCL Non-live}
      & \multicolumn{2}{c}{HotpotQA}
      & \multicolumn{2}{c}{2Wiki}
      & \multicolumn{2}{c}{MuSiQue}\\
    \cmidrule(lr){2-3}\cmidrule(lr){4-5}\cmidrule(lr){6-7}\cmidrule(lr){8-9}\cmidrule(l){10-11}
      & {\fontsize{8}{9.5}\selectfont Acc.\,$\uparrow$} & {\fontsize{8}{9.5}\selectfont FLOPs\,$\downarrow$}
      & {\fontsize{8}{9.5}\selectfont Acc.\,$\uparrow$} & {\fontsize{8}{9.5}\selectfont FLOPs\,$\downarrow$}
      & {\fontsize{8}{9.5}\selectfont Acc.\,$\uparrow$} & {\fontsize{8}{9.5}\selectfont FLOPs\,$\downarrow$}
      & {\fontsize{8}{9.5}\selectfont Acc.\,$\uparrow$} & {\fontsize{8}{9.5}\selectfont FLOPs\,$\downarrow$}
      & {\fontsize{8}{9.5}\selectfont Acc.\,$\uparrow$} & {\fontsize{8}{9.5}\selectfont FLOPs\,$\downarrow$} \\
    \midrule
    w/o flux policy & 70.07 & 26.95 & 64.46 & 17.39 & 47.60 & 149.71 & 66.70 & 162.46 & 19.60 & 162.93 \\
    w/o conf. align. & 69.63 & 23.11 & 64.46 & 14.55 & \textbf{48.90} & 145.49 & \textbf{68.90} & 166.48 & 19.10 & 167.32 \\
    \textbf{Controller (Ours)} & \textbf{70.96} & \textbf{18.48} & \textbf{65.47} & \textbf{12.51} & 46.60 & \textbf{131.91} & 67.90 & \textbf{147.68} & \textbf{20.30} & \textbf{142.83} \\
    \bottomrule
  \end{tabular*}
  \caption{Component ablations of the metacognitive controller.
  Each ablation disables one component of the full controller.
  Conf. align. denotes the confidence alignment policy.
  Accuracy (\%) and FLOPs ($10^{12}$) follow Table~\ref{tab:backbone-em}.}
  \label{tab:component-ablation}
\end{table*}

\paragraph{Implementation and evaluation metrics.}
To support a fair comparison with task-adapted baselines, we first adapt LLaDA-8B-Instruct~\citep{nie2025llada} to the target tasks through supervised fine-tuning (SFT). For the controller, we simply use a two-layer Transformer encoder. We train the controller on eight NVIDIA A100 GPUs for 40 rounds per stage.

\textbf{Task accuracy.}
For BFCL v3, we use the official evaluator to measure tool-call correctness.
For HotpotQA, 2WikiMultiHopQA, and MuSiQue, we report
\textbf{cover exact match (CEM)}, following prior evaluations of
agents~\citep{dllmsearcher,song2025r1searcher}.
CEM assesses whether the ground-truth answer is included in the predicted answer.
CEM scores are rounded to the nearest 0.1 percentage point.

\textbf{Computational cost.}
Floating-point operations (\textbf{FLOPs}) measure the amount of arithmetic
computation required for inference.
We count the major matrix operations of the backbone and controller
over each complete trajectory and report average FLOPs in units of $10^{12}$ operations.

\subsection{Main Results}
\label{sec:exp-main}
\label{sec:exp-search}

Table~\ref{tab:backbone-em} summarizes the main results across five benchmarks.
Our method delivers competitive task accuracy at the lowest computational cost compared with existing baselines.
Notably, on BFCL Live, HotpotQA, and 2Wiki, it outperforms all baselines in both accuracy and inference FLOPs.

\paragraph{Comparison with general diffusion models.}
Our method builds on LLaDA-8B-Instruct.
Compared with the larger-scale LLaDA2.0-mini (16B MoE), it achieves higher accuracy on four benchmarks while reducing inference FLOPs across all five benchmarks.
Although LLaDA2.0-mini achieves higher accuracy on BFCL Non-live, our method still reduces inference FLOPs by 11.4\% on this benchmark.

\paragraph{Comparison with task-specialized models.}
Our method consistently outperforms LLaDA-8B-Tools and DLLM-Searcher in both task accuracy and computational efficiency across all five benchmarks.
On BFCL Non-live and Live, it surpasses LLaDA-8B-Tools by 28.20 and 20.96 percentage points in accuracy, respectively.
The advantage over DLLM-Searcher extends across all three multi-hop QA benchmarks, with accuracy gains of 9.50, 17.50, and 2.50 percentage points on HotpotQA, 2Wiki, and MuSiQue, respectively.
These improvements are accompanied by an 88.3--89.7\% reduction in inference FLOPs across these three benchmarks.

\paragraph{Comparison with decoding acceleration baselines.}
Compared with Fast-dLLM, our method reduces inference FLOPs by 56.6--77.2\% across all five benchmarks while maintaining competitive task accuracy.
On BFCL Non-live and Live, it also achieves accuracy gains of 12.81 and 13.50 percentage points, respectively, demonstrating improvements in both accuracy and computational efficiency.

\paragraph{Effectiveness of metacognitive control.}
Compared with LLaDA-8B (SFT), our method reduces inference FLOPs
by 24.6--43.1\% across all five benchmarks with identical backbone
weights and inference settings, differing only in the use of
the metacognitive controller.
Task accuracy improves on four benchmarks and remains comparable
on MuSiQue, demonstrating the computational benefits of
metacognitive control.

\subsection{Ablation Study}
\label{sec:exp-ablation}
\label{sec:exp-analysis}

We examine the contributions of the two decoding components in Table~\ref{tab:component-ablation}.
We disable the flux decoding policy and the confidence alignment policy individually. All configurations follow our method's inference setup, including prefix caching and confidence-threshold decoding.

\paragraph{Contributions of individual components.}
As shown in Table~\ref{tab:component-ablation}, removing either
policy increases inference FLOPs across all five benchmarks,
with only modest changes in task accuracy.
Specifically, disabling the flux decoding policy and the confidence
alignment policy individually increases FLOPs by an average of
24.5\% and 16.3\%, respectively, relative to the full controller.
These results demonstrate that both policies contribute to the
controller's efficiency gains while preserving comparable task accuracy.

\begin{figure}[t]
  \centering
  \resizebox{\linewidth}{!}{\begingroup
\sffamily
\pgfplotsset{compat=1.18}
\definecolor{front}{HTML}{245C91}
\definecolor{muted}{HTML}{BFC6CD}
\begin{tikzpicture}
\begin{groupplot}[group style={group size=1 by 2,vertical sep=1.8cm},width=8.3cm,height=5.2cm,
xlabel={FLOPs reduction (\%) $\uparrow$},
axis lines*=left, axis line style={black!40}, tick style={black!30},
ymajorgrids=true,grid style={black!8},tick label style={font=\small},
label style={font=\small},title style={font=\normalsize},scaled ticks=false,
legend style={font=\scriptsize,draw=none,fill=white,at={(0.025,0.025)},anchor=south west},
]
\nextgroupplot[title={BFCL},ylabel={Mean accuracy (\%) $\uparrow$},xmin=26,xmax=39,ymin=59.8,ymax=68.5,xtick={28,30,32,34,36,38}]
\addplot[only marks,mark=*,mark size=2pt,color=muted] coordinates {(30.654512779,66.369047619) (31.443351547,65.476190476) (31.871594644,63.888888889) (34.121900962,64.781746032) (34.207070309,64.087301587) (36.451468667,63.888888889) (35.656920362,64.087301587) (36.708890268,62.500000000) (35.316762152,63.591269841) (33.933358250,62.500000000) (36.075027700,63.392857143) (35.978540820,61.805555556) (34.126116757,63.888888889) (37.854069659,60.714285714) (36.291411130,63.194444444) (35.818478487,64.980158730) (27.627762117,67.261904762)};
\addlegendentry{Checkpoints}
\addplot[color=front,line width=1pt,mark=*,mark size=2.8pt,mark options={fill=front,draw=white,line width=0.5pt}] coordinates {(37.854069659,60.714285714) (36.708890268,62.500000000) (36.451468667,63.888888889) (35.818478487,64.980158730) (31.443351547,65.476190476) (30.654512779,66.369047619) (27.627762117,67.261904762)};
\addlegendentry{Empirical frontier}
\draw[-{Stealth[length=4pt]},black!55,line width=0.65pt] (rel axis cs:0.70,0.79) -- (rel axis cs:0.90,0.93);
\node[font=\scriptsize,text=black!60,anchor=south east] at (rel axis cs:0.82,0.91) {Better};
\nextgroupplot[title={Multi-hop Dataset},ylabel={Mean CEM (\%) $\uparrow$},xmin=18.5,xmax=23.5,ymin=43.25,ymax=45,xtick={19,20,21,22,23},ytick={43.5,44,44.5,45}]
\addplot[only marks,mark=*,mark size=2pt,color=muted] coordinates {(19.321319368,44.555555556) (21.839221441,44.000000000) (21.496571554,44.000000000) (22.435500706,43.555555556) (22.985135106,43.888888889) (22.075921708,43.777777778) (22.430776499,43.777777778) (20.904258736,44.222222222) (20.772546312,43.888888889)};
\addlegendentry{Checkpoints}
\addplot[color=front,line width=1pt,mark=*,mark size=2.8pt,mark options={fill=front,draw=white,line width=0.5pt}] coordinates {(22.985135106,43.888888889) (21.839221441,44.000000000) (20.904258736,44.222222222) (19.321319368,44.555555556)};
\addlegendentry{Empirical frontier}
\draw[-{Stealth[length=4pt]},black!55,line width=0.65pt] (rel axis cs:0.70,0.79) -- (rel axis cs:0.90,0.93);
\node[font=\scriptsize,text=black!60,anchor=south east] at (rel axis cs:0.82,0.91) {Better};
\end{groupplot}\end{tikzpicture}

\endgroup}
  \caption{Accuracy versus FLOPs reduction on BFCL and multi-hop tasks. Gray points denote evaluated checkpoints; blue points and lines highlight their empirical Pareto frontier. The upper-right direction is better.}
  \label{fig:accuracy-cost}
\end{figure}
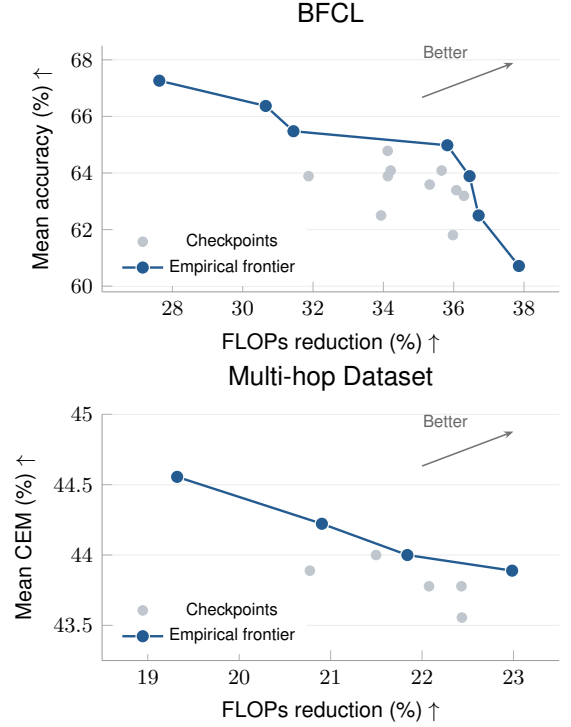

\paragraph{Accuracy--efficiency trade-off.}
Figure~\ref{fig:accuracy-cost} plots task accuracy against FLOPs reduction on BFCL and multi-hop tasks for several controller checkpoints. For each checkpoint on the highlighted Pareto frontier, none of the other checkpoints improves either metric without worsening the other.

\subsection{Results on Different Backbones}
\label{sec:exp-backbones}

We further evaluate our method on two different diffusion language model backbones:
LLaDA-8B-Instruct~\citep{nie2025llada} and Dream-7B~\citep{ye2025dream}.
We train the metacognitive controller for each backbone separately.
Table~\ref{tab:backbone-effect} shows substantial reductions in inference FLOPs on both backbones while retaining competitive success rates, demonstrating the effectiveness of our method across different diffusion backbones.

\begin{table}[t]
  \centering
  \small
  \setlength{\tabcolsep}{2pt}
  \renewcommand{\arraystretch}{1.1}
  \begin{tabular*}{\columnwidth}{@{\extracolsep{\fill}}lrrrr@{}}
    \toprule
    \multirow{2}{*}{Dataset}
      & \multicolumn{2}{c}{LLaDA}
      & \multicolumn{2}{c}{Dream} \\
    \cmidrule(lr){2-3}\cmidrule(l){4-5}
      & $\Delta$Acc. & FLOPs$\downarrow$
      & $\Delta$Acc. & FLOPs$\downarrow$ \\
    \midrule
    BFCL Live     & $+4.44$ & 43.1\% & $+0.00$  & 17.2\% \\
    BFCL Non-live & $+1.87$ & 35.1\% & $+1.72$ & 29.0\% \\
    HotpotQA      & $+0.90$ & 24.6\% & $+1.10$ & 52.9\% \\
    2Wiki         & $+0.80$ & 27.8\% & $-0.60$ & 58.0\% \\
    MuSiQue       & $-0.40$ & 27.6\% & $-2.20$ & 67.5\% \\
    \bottomrule
  \end{tabular*}
  \caption{Effect of the metacognitive controller on LLaDA and Dream.
  $\Delta$Acc. reports the change in accuracy (percentage points), and
  FLOPs$\downarrow$ reports the reduction in inference FLOPs (\%),
  relative to the same backbone without the controller.
  LLaDA results compare Ours with LLaDA-8B (SFT) in
  Table~\ref{tab:backbone-em}.}
  \label{tab:backbone-effect}
\end{table}

\begin{figure}[!t]
  \centering
  \includegraphics[width=\columnwidth]{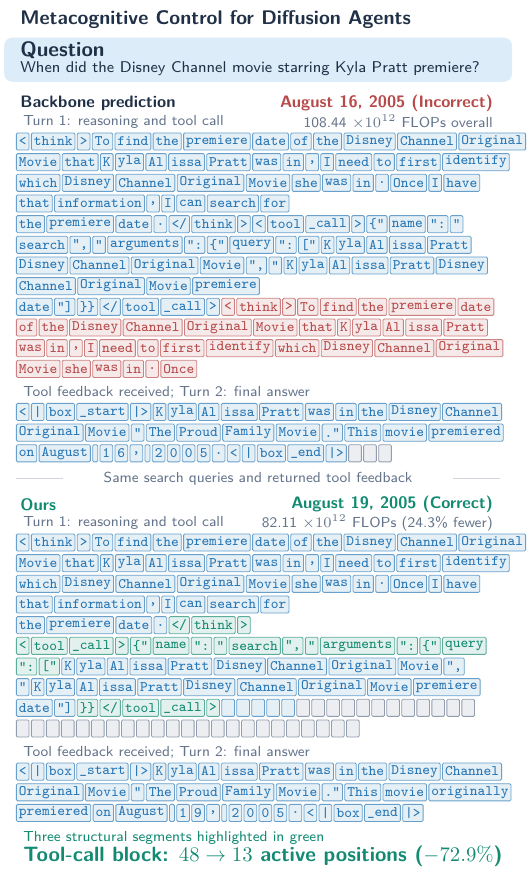}
  \caption{An interaction comparison using the same backbone and search queries.
  Green highlights the tokens in three structural segments; Red marks post-call generation; gray denotes padding. Blank blue cells have no saved token text. Flux decoding reduces the active range (gray).
  With the metacognitive controller, the agent uses 24.3\% fewer FLOPs over the complete interaction and answers correctly in this example. }
  \label{fig:case-study}
\end{figure}

\subsection{Case Study}
\label{sec:exp-case-study}

Figure~\ref{fig:case-study} compares two complete interactions using the same diffusion backbone and search queries.
Both agents issue a tool call in the first turn and generate a final answer in the second.
With backbone prediction alone, generation continues beyond the tool-call boundary and repeats the preceding reasoning.
These 40 positions are discarded by the environment: they neither change the search queries nor contribute to the next turn's interaction context.
The decoder therefore spends computation on this redundant content.

The metacognitive controller coordinates structural and token acceptance through the confidence alignment policy and adjusts the active computation range through flux decoding.
The figure highlights three structural segments in green: the thinking close tag, the search-call prefix, and the tool-call closing sequence.
In the final tool-call block, flux decoding retains 13 of 48 positions, reducing the active range within this block by 72.9\%.
After receiving the same tool feedback, the agent with our controller gives the correct premiere date, August 19, 2005, whereas the agent relying on backbone prediction alone answers August 16, 2005.
The complete interaction uses $82.11\times10^{12}$ FLOPs, compared with $108.44\times10^{12}$ with backbone prediction alone, a reduction of 24.3\%.

\section{Conclusion}
\label{sec:conclusion}

We presented a metacognitive controller designed to accelerate diffusion language agents through environment-aware generation. Rather than processing tokens uniformly in fixed blocks, our controller dynamically regulates decoding via a cascaded policy: confidence alignment decoding finalizes predictable structural tokens efficiently, while flux decoding shrinks the active computation range to eliminate redundant processing. Empirical results across diverse tool-calling and multi-hop reasoning benchmarks show that our method substantially reduces inference FLOPs without compromising task accuracy. This approach demonstrates that adaptively aligning computation with agent interaction states is a highly effective strategy for scaling diffusion agents.

\section*{Limitations}
\label{sec:limitations}

Adapting the controller to evolving diffusion models remains an important direction for future work.

\bibliography{custom}

\end{document}